# Machine Learning Framework for Thrombosis Risk Prediction in Rotary Blood Pumps

Christopher Blum[a], Michael Neidlin[a, *]

Affiliation:

a) Cardiovascular Engineering, Applied Medical Engineering, RWTH Aachen University, Aachen, Germany

*Correspondence:

Name: Michael Neidlin

Address: Institute of Applied Medical Engineering, Forckenbeckstr. 55

52074 Aachen, Germany

Email address: michael.neidlin@gmail.com

Funding:

This research received no specific grant from any funding agency in the public, private, or not-for-profit sectors

Conflict of interest:

All of the authors have nothing to disclose.

Authors' contributions:

All authors contributed to the study conception and design. CB developed the numerical model, performed the simulations, gathered, analyzed and discussed the results. MN was involved in the analysis and discussion of the results. MN supervised the project. CB wrote the manuscript based on the input of all co-authors. All co-authors read and approved the final version of the manuscript.

# Abstract:

Thrombosis in rotary blood pumps arises from complex flow conditions that remain difficult to translate into reliable and interpretable risk predictions using existing computational models. This limitation reflects an incomplete understanding of how specific flow features contribute to thrombus initiation and growth.

This study introduces an interpretable machine learning framework for spatial thrombosis assessment based directly on computational fluid dynamics-derived flow features. A logistic regression (LR) model combined with a structured feature-selection pipeline is used to derive a compact and physically interpretable feature set, including nonlinear feature combinations. The framework is trained using spatial risk patterns from a validated, macro-scale thrombosis model for two representative scenarios. The model reproduces the labeled risk distributions and identifies distinct sets of flow features associated with increased thrombosis risk. When applied to a centrifugal pump, despite training on a single axial pump operating point, the model predicts plausible thrombosis-prone regions.

These results show that interpretable machine learning can link local flow features to thrombosis risk while remaining computationally efficient and mechanistically transparent. The low computational cost enables rapid thrombogenicity screening without repeated or costly simulations. The proposed framework complements physics-based thrombosis modeling and provides a methodological basis for integrating interpretable machine learning into CFD-driven thrombosis analysis and device design workflows.

# 1. Introduction

Thrombosis remains a major challenge in the design and clinical operation of rotary blood pumps [1–3]. These devices expose blood to complex three-dimensional flow conditions in which combinations of supraphysiological shear stress, recirculation and turbulent structures can activate platelets and initiate thrombus formation [4–6]. While computational fluid dynamics (CFD) is widely used to characterize such environments, translating detailed flow fields into reliable predictions of thrombosis risk continues to be difficult [7]. Traditional blood damage models, including shear-based platelet activation models [8, 9] and macroscopic thrombosis formulations [10–12], have contributed substantially to the understanding of flow-induced clotting mechanisms. However, these approaches often require high computational effort, fall short on capturing all relevant mechanisms of thrombosis or rely on thresholding procedures, which provide limited insight into the specific flow features responsible for thrombus initiation and growth. In particular, they typically focus on a small set of commonly used flow features, such as velocities or shear rates, despite the fact that CFD simulations provide a much richer set of local flow descriptors whose relationship to thrombosis remains largely unexplored. As a consequence of these combined limitations, the practical application of such models in iterative device development or therapy planning remains limited.

In parallel, machine learning has achieved substantial success across biomedical engineering, particularly in image-based applications [13, 14] and in the prediction of clinical outcomes from large-scale patient datasets [15], where the availability of extensive training data has enabled highly accurate and robust models. These approaches are now well established and widely accepted as powerful tools for medical analysis and decision support. In the context of blood-contacting devices, data-driven methods have also begun to emerge, including surrogate modeling of flow features across different blood pump operating points [16] and machine learning assisted multiscale thrombosis modeling aimed at accelerating computationally expensive simulations [17, 18]. These studies highlight the potential of machine learning to complement CFD-based investigations and to reduce computational cost in complex hemodynamic modeling pipelines. However, existing approaches in this domain are typically not built directly on local flow features alone or rely on hybrid modeling strategies in which machine learning is embedded within multiscale frameworks. As a result, key parts of the prediction process remain non-interpretable, and the direct link between CFD-derived flow features and thrombosis risk is obscured. This represents an important limitation for medical device applications, since regulatory approval frameworks emphasize the need for physically interpretable and mechanistically grounded modeling approaches [19].

Interpretable machine learning offers an alternative pathway by providing fast and reproducible predictions while maintaining explicit connections between model outputs and the underlying flow features. Such an approach is particularly attractive for thrombosis assessment, where mechanistic insight into flow–platelet interactions is relevant for both device optimization and further improvement of modeling techniques. Despite this potential, interpretable models have rarely been used to identify spatially resolved thrombosis risk directly from CFD simulations, and no established framework exists for linking flow features to thrombus-prone regions in rotary blood pumps.

The aim of this study is to address this gap by developing and evaluating an interpretable machine learning framework for spatial thrombosis prediction based on CFD-derived flow features. The approach integrates a logistic regression (LR) model with a structured feature-selection pipeline to produce a compact and physically grounded set of predictors. The study evaluates the model's ability to reproduce predictions of macroscopic thrombosis model risk patterns, to identify the flow features that characterize the main thrombosis mechanisms, and to demonstrate preliminary generalization across pump types and operating conditions.

## 2. Methods

### 2.1 Reference thrombosis model

The present work builds upon a previously published macroscopic thrombosis model [20]. The model couples steady-state CFD with several scalar transport equations that describe the mechanical and biochemical activation of platelets. It has been validated for the HeartMate II rotary blood pump and provides spatially resolved predictions of regions with elevated thrombosis risk. Details of the thrombosis model implementation and the CFD setup are available in the original publication [20].

For the current study, the thrombosis-model outputs were used as ground-truth labels for the machine-learning framework. The activated platelet (AP) field was scaled in accordance with the original model formulation by expressing the per-mille increase relative to an initial concentration of $2.5 \times 10^{13} m^{-3}$. This normalization leads to a scaled activated platelets (sAP) variable that ensures consistency with the interpretation of the AP magnitude used in the earlier work.

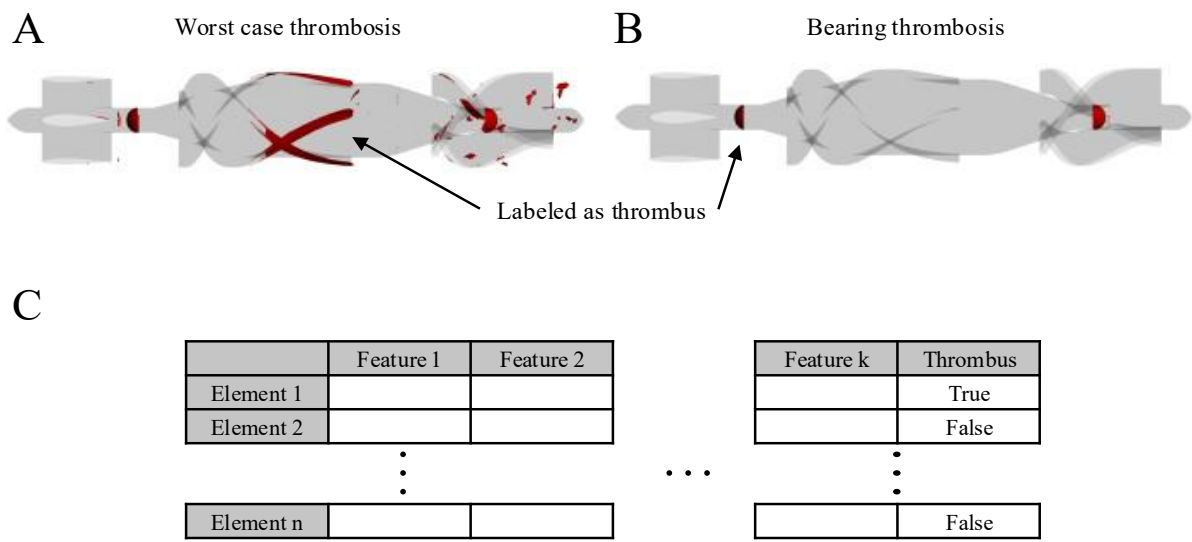

*Figure 1: Overview of thrombosis label sets and feature-based dataset construction. Worst-case and bearing thrombosis scenarios are defined by applying different thresholds to the scaled activated platelet (sAP) of 6 and 12, depicted in A and B respectively. This results in spatially distributed and localized thrombus labels, which can be transformed to a table of flow features together with a binary thrombosis label (C), yielding a dataset of approximately n=4 million elements with k=178 candidate flow features and a thrombus label rate of about 2 % for A and 0.7 % for B.*

Because the thrombosis model identifies regions of increased thrombosis risk rather than predicting absolute thrombus formation, two labeling scenarios were defined. For the operating point of 9000 rpm and 2 L/min, all computational cells with sAP values above 6 were labeled as thrombus in the first scenario, representing a broad, worst-case condition with strong thrombus potential (Figure 1A). In the second scenario, only cells exceeding a sAP value of 12 were labeled as thrombus, capturing exclusively the most elevated sAP regions across the front and rear bearing, as depicted in Figure 1B. All remaining cells were labeled as non-thrombus. For each scenario, all computational cells were exported in a tabular structure containing the local flow variables in scalar from together with the binary "Thrombus" label, which categorizes each cell into thrombus or non-thrombus. These datasets served as the basis for training and evaluating the machine learning model.

**2.2 Machine learning pipeline**

A structured, multi-stage machine learning pipeline was implemented to identify a compact and physically interpretable set of flow-derived features for thrombus classification. The overall workflow is summarized in Figure 2 and comprises four sequential stages:

1. initial reduction of the full feature set, including redundancy removal and coefficient-based importance screening at the 1% level
2. training of a baseline LR model on the resulting reduced set of 14 features
3. construction of engineered nonlinear and interaction features, expanding these 14 features to 126 engineered variables
4. recursive feature elimination based on leave-one-feature-out cross-validation.

Each stage was designed to systematically reduce dimensionality, improve model expressiveness, address the strong class imbalance in the dataset, and ensure that only features with demonstrable predictive relevance were retained.

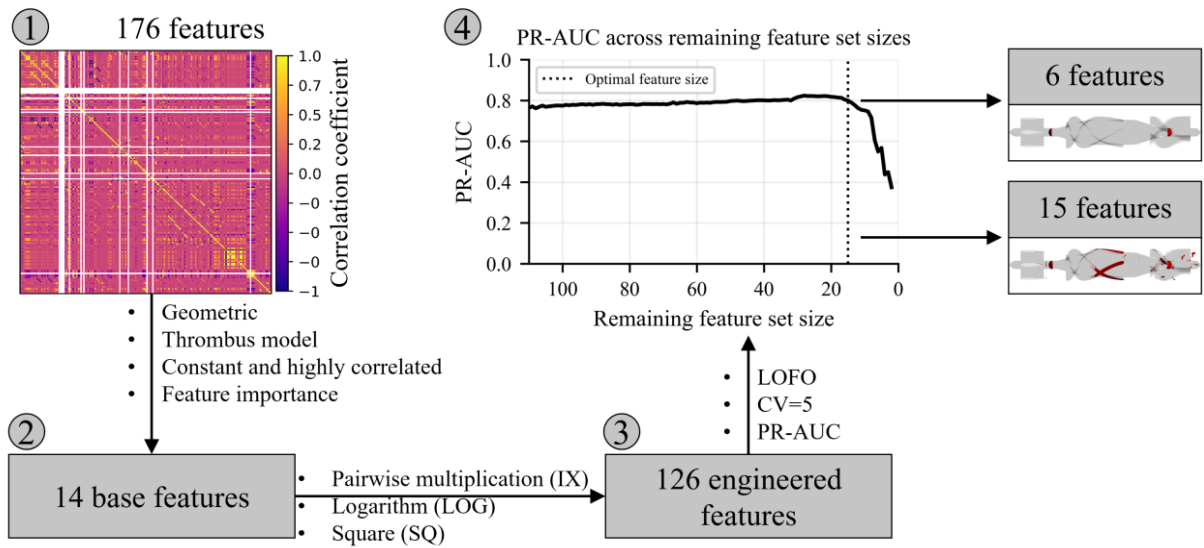

*Figure 2: Multi-stage feature-processing pipeline for thrombosis classification. Step 1 reduces the initial feature set by removing redundant and non-informative features. Step 2 trains a baseline logistic regression (LR) model on the reduced feature set. Step 3 constructs nonlinear and interaction features to expand the feature space. Step 4 applies recursive feature elimination using leave-one-feature-out (LOFO) cross-validation (CV) to identify a compact and interpretable final feature set based on precision–recall area under the curve (PR-AUC).*

The dataset exhibited substantial class imbalance, with thrombus-labeled samples representing only a small minority of 2 % or even 0.7% in the worst case and bearing case, respectively. To mitigate this, all models employed class-balanced LR, assigning weights inversely proportional to class frequencies. Data were split using stratified sampling to preserve the natural thrombus-to-non-thrombus ratio. First, 20% of the data were held out as a final test set. The remaining 80% were divided into 80% training and 20% validation subsets. All feature selection and model refinement steps were conducted exclusively using the training data; the validation set served to monitor model performance.

Initial reduction of the full feature set began with 178 flow-derived features. All variables containing geometric information, features originating directly from the thrombosis model, constant features, and highly correlated pairs were removed. For correlated pairs, one representative variable was retained. To provide an initial ranking of the remaining variables, a LR model with L2 regularization was fitted using all retained features. The probability P of thrombus formation was modeled by LR as:

$$P(y = 1 \mid \mathbf{x}) = \sigma(z) \quad , \quad z = \beta_0 + \sum_{j=1}^{d} \beta_j x_j \tag{1}$$

where $\mathbf{x} = (x_1, \dots, x_d)$ is the feature vector and $\beta_j$ are the learned coefficients. The logistic function is defined as:

$$\sigma(z) = \frac{1}{1 + e^{-z}} \tag{2}$$

Here, $z$ is the linear predictor, i.e. the weighted sum of the standardized input features plus the intercept.

Feature importance was quantified by the absolute coefficient magnitude

$$I_j = |\beta_j| \tag{3}$$

and normalized by the sum of importance values across all $d$ features as:

$$\tilde{I}_j = \frac{I_j}{\sum_{k=1}^{d} I_k} \tag{4}$$

To conclude the first step of removing all unnecessary features that did not contribute to the prediction task, all features with normalized importance below 1% were discarded. This produced a reduced base feature set of 14 base features.

A baseline LR model using these 14 base features was trained to assess linear separability. Despite appropriate class weighting, the baseline model exhibited limited discriminative capacity. Model performance was evaluated using precision–recall (PR) curves, which provide a more informative assessment than ROC-AUC for imbalanced datasets by focusing on performance for the minority class. For any probability threshold, precision and recall are defined as:

$$\text{Precision} = \frac{TP}{TP + FP} \quad , \quad \text{Recall} = \frac{TP}{TP + FN} \tag{5}$$

where TP, FP, and FN denote true positives, false positives, and false negatives. The area under the PR curve (PR-AUC) summarizes performance across all thresholds and directly reflects the model's ability to detect thrombus samples while limiting false positives. The PR-AUC for the baseline feature set was low, indicating the need for nonlinear transformations.

To enrich model expressiveness, the 14 base features were therefore expanded into 126

engineered features in the third step of the pipeline. For each feature $x_j$, the following nonlinear transformations were generated:

- squared terms $x_j^2$ (abbreviated as SQ)
- logarithmic transformations $\log(x_j + c)$ with a constant shift for numerical stability (abbreviated as LOG)
- pairwise interaction terms $x_j x_k$ (abbreviated as IX)

Together with the original baseline features these engineered features allowed the model to capture nonlinear and interaction effects that are plausible in shear-driven thrombosis processes. The engineered feature matrix was standardized on the training data.

Because the engineered feature space of 126 features prohibits an interpretable model outcome, recursive feature elimination was performed using leave-one-feature-out (LOFO) cross-validation (CV). A five-fold stratified CV scheme was applied. For each iteration, a baseline cross-validated PR-AUC was computed using all currently retained engineered features. Then each feature was removed in turn, a model was refitted, and the PR-AUC difference

$$\Delta_j = \text{PR-AUC}_{\text{baseline}} - \text{PR-AUC}_{\text{without } f_j} \tag{6}$$

was recorded. LOFO importances were smoothed over 5 consecutive LOFO iterations to reduce variability. At each iteration, the feature with the lowest smoothed LOFO importance was removed. As depicted by the dashed line in the fourth step of Figure 2, this elimination procedure continued until further removal produced a consistent decline in PR-AUC, ensuring that only features with demonstrable predictive relevance were retained.

The procedure resulted in a compact and interpretable subset of engineered features that preserved high PR-AUC performance despite the severe class imbalance. This final feature set served as the basis for all subsequent model evaluations and analyses.

The complete Python code for the multi-stage machine learning pipeline, including the final Ansys CFX expression of equation 2, is available to the scientific community at https://doi.org/10.5281/zenodo.17901009, providing an open-source basis for further development and reproducibility.

# 3. Results

This chapter presents the performance and behavior of the data-driven thrombosis prediction framework. The analysis is structured to first evaluate how accurately the LR model reproduces the thrombosis-risk patterns obtained from the underlying macroscopic thrombosis model. The chapter then examines which flow-derived features are most relevant for the model's predictions, followed by an assessment of how well the trained model generalizes to a different pump type and operating condition.

### 3.1 Prediction accuracy

Figure 3 illustrates the LR model predictions for two training scenarios based on the HeartMate II geometry at 9000 rpm and 2 L/min. Since the underlying thrombosis model does not directly predict thrombus formation zones but instead identifies regions of elevated risk based on thresholds of sAP, two scenarios were defined. Scenario A ("worst case thrombosis") applies a lower threshold of sAP = 6, resulting in a more widespread distribution of labeled high-risk regions throughout the pump. Scenario B ("bearing thrombosis") applies a higher threshold of sAP = 12, producing high-risk labels exclusively in the bearing region. For both scenarios, the LR model shows good agreement with the training data. The predicted thrombosis probability is indicated using a white-to-red color scale, with red corresponding to the highest probability. In Scenario A, the model reproduces the complex spatial patterns of high-risk regions along the rotor blades and bearing sections but misses some of the regions in the outlet section. In Scenario B, the model accurately highlights elevated probabilities around the bearings, consistent with the localized nature of the training data. These results demonstrate that the LR model captures the distinct spatial risk distributions imposed by the two labeling thresholds.

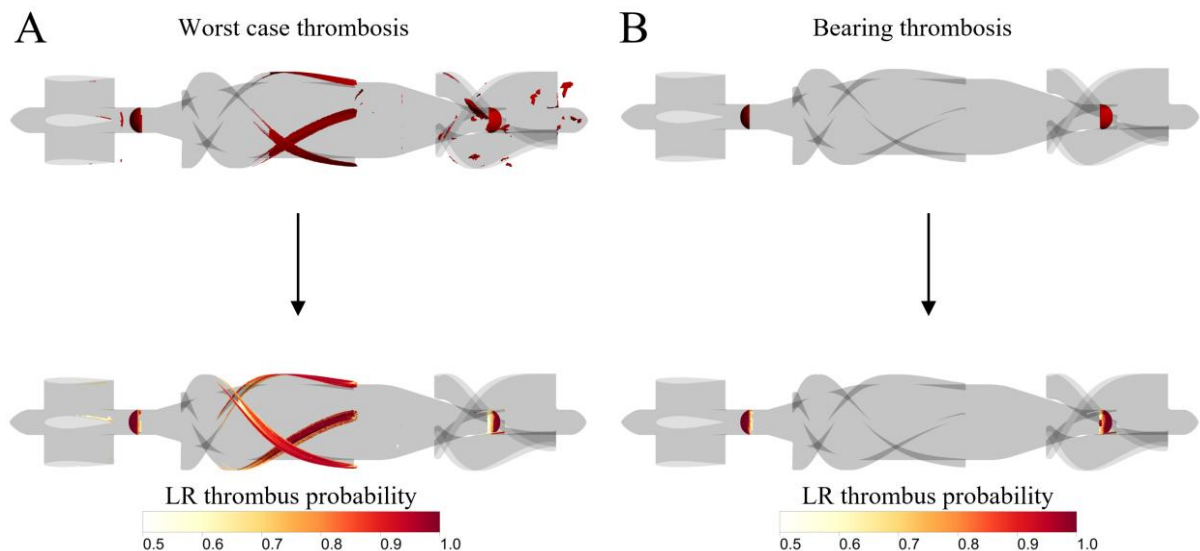

*Figure 3: Logistic regression (LR) model predictions (bottom) for two labeling scenarios at 9000 rpm and 2 L/min (top). In A, the worst-case scenario uses an sAP threshold of 6 and results in widespread high-risk regions. In B, the bearing-thrombosis scenario uses an sAP threshold of 12 and confines high-risk regions to the bearing. Predicted probabilities range from white to red and match the training data well. SQ denotes squared terms, LOG logarithmic transformations, and IX pairwise interaction terms.*

To understand how the model achieves this level of predictive performance, the next section examines which features contribute most strongly to the LR predictions.

### 3.2 Feature importance

Figure 4 presents the permutation feature importance for the final LR models corresponding to the two training scenarios. In Scenario A, the model relies on 15 features, with several contributing similarly to the prediction. Except for the variable Velocity.Invariant Q, all highly ranked features are engineered combinations of flow quantities. The most influential features are dominated by variables linked to shear strain rate, vortical structures, and pressure-related terms, indicating that thrombosis risk in the worst-case scenario is strongly associated with these flow characteristics. In contrast, Scenario B identifies only five features as important for accurately predicting thrombosis in the bearing region.

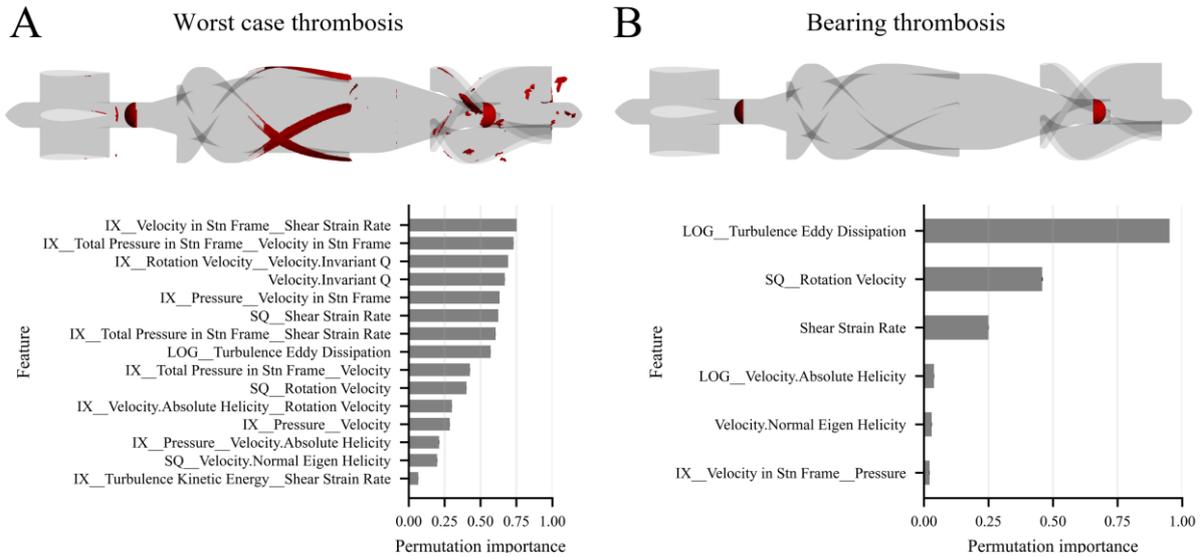

*Figure 4: Permutation feature importance for the LR models in both labeling scenarios. In A, 15 features contribute to the worst-case predictions, mostly engineered combinations related to shear strain rate, vortical structures, and pressure. In B, the bearing-thrombosis scenario requires only 5 features, with turbulence eddy dissipation, rotational velocity, and shear strain rate being most influential.*

The three most influential features are turbulence eddy dissipation, rotational velocity, and shear strain rate, which the permutation analysis highlights as the key drivers of the model's decisions. With this, the model offers insight into the specific flow mechanisms that distinguish the thrombosis-prone bearing region from other parts of the pump.

### 3.3 Generalization across pump types

Figure 5 shows the transfer of the HeartMate II–trained LR model to a centrifugal blood pump (FDA benchmark [21]) operated at 3500 rpm and 6 L/min.

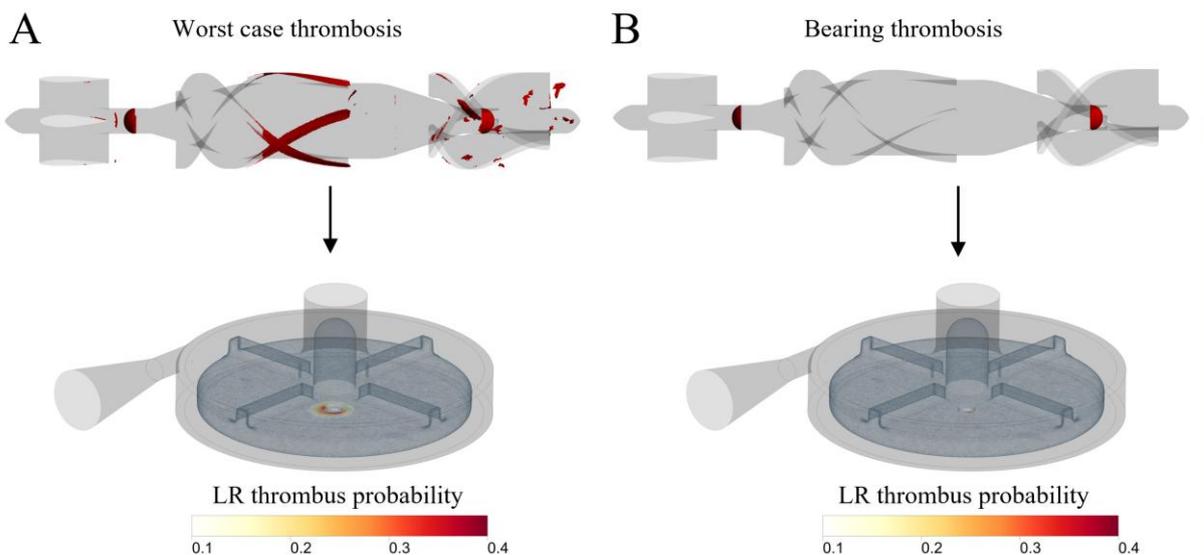

*Figure 5: Application of the HeartMate II–trained LR model to the FDA benchmark centrifugal pump at 3500 rpm and 6 L/min. In both A and B, the model predicts elevated thrombosis probability near the bottom of the impeller-eye region, consistent with areas typically prone to thrombus formation in centrifugal pumps.*

Despite being trained exclusively on a single operating point of an axial pump, the model produces reasonable thrombosis-probability patterns in a geometrically and hydraulically different pump. In both training scenarios, the transferred model identifies increased probability near the bottom of the impeller eye, a region associated with prolonged residence times that has been suggested as a potential thrombosis-prone area in centrifugal pumps [22].

## 4. Discussion

The aim of this study was to develop and evaluate an interpretable machine learning framework for thrombosis prediction in rotary blood pumps that operates directly on CFD-derived flow features. Using a LR model with nonlinear feature combinations, the framework reproduces the spatial risk patterns obtained from a validated macroscopic thrombosis model and identifies flow variables associated with different thrombosis scenarios. Despite being trained on a single operating point of an axial device, the model shows preliminary transferability when applied to a centrifugal pump, suggesting that it captures meaningful relationships between local flow features and thrombosis risk. Together, these results demonstrate that interpretable machine learning can complement CFD-based thrombosis modeling by providing spatial predictions with mechanistic insight in a computationally efficient manner.

A closer examination of the prediction accuracy provides further insight into the behavior of the framework. Because the underlying macroscopic thrombosis model provides risk information by thresholding scaled activated platelets rather than predicting actual thrombus morphology, two training scenarios were defined to represent broad and localized risk patterns. Linear combinations of flow features were not sufficient to reproduce the worst-case scenario, whereas feature sets containing non-linear combinations enabled the model to recover both the worst-case distributed and the bearing-focused risk regions. This is consistent with literature indicating that thrombosis in rotary blood pumps is influenced by combined effects of shear exposure, residence time, vortical transport, and recirculation zones rather than a single dominant flow mechanism [22]. However, an important limitation of the present analysis is the reliance on threshold-based training labels, which do not represent experimentally verified thrombus locations. This limits the extent to which the results can be interpreted in a clinically meaningful way and highlights the ongoing need for spatially resolved thrombosis data. It also emphasizes the importance of understanding how uncertainty in threshold selection may influence model performance, since different cutoff values can substantially alter the spatial distribution of the training labels.

The feature importance analysis expands this perspective by showing that the two thrombosis scenarios differ not only in spatial extent but also in the mechanisms that characterize them. The worst-case scenario involved a broad range of influential flow features, which reflects the complex interplay of shear, vortical structures, pressure gradients, and rotational components. In contrast, the bearing-focused scenario was dominated by turbulence eddy dissipation, rotational velocity, and shear strain rate. This difference illustrates the strength of interpretable machine learning, which not only predicts where thrombosis may occur but also clarifies which flow mechanisms are characteristic of different risk patterns. A key strength of the interpretable machine learning framework arises from the structured and generalizable feature-selection pipeline, which combines redundancy removal, coefficient-based screening, engineered nonlinear features, and LOFO selection to derive a compact and physically interpretable feature set. This ensures that the final features remain mechanistically meaningful while maintaining good model performance. Although the presented framework provides a useful complement to existing thrombosis models, it remains limited by its reliance on Eulerian and instantaneous flow quantities. Since thrombosis develops through a convective and time-integrated process, the use of residence time, particle activation history, or similar Lagrangian descriptors would provide a more complete representation of the underlying mechanisms.

The ability of the model to generate plausible predictions in a centrifugal pump, despite being trained on an axial flow pump setup, further supports the methodological potential of the framework. Both axial flow pump training scenarios resulted in increased predicted risk near the bottom of the impeller eye in the centrifugal pump, which is suggested to be a thrombus-prone regions in centrifugal devices [22]. This observation does not imply that the current model is generally applicable across all pump types but rather suggests that the method is capable of learning flow-feature relationships that extend beyond the original training case. More comprehensive validation, including multiple pump geometries, operating conditions, and experimentally confirmed thrombus patterns, is needed to assess the scope and reliability of this generalization. At present, the lack of real-world spatial thrombosis data in medical devices remains a fundamental constraint. Until such data become accessible, macroscopic thrombosis models will continue to provide the most practical source of training information, despite their dependence on user-defined thresholds. The framework should therefore be viewed as a complement to physics-based thrombosis modeling [11, 23, 24], as its low computational cost, enabled by the simplicity of the LR model and the compact feature set, allows fast screening of thrombogenicity, which is particularly useful for iterative design workflows.

Even with these limitations, the presented framework opens several avenues for future development. A first step is the incorporation of convective and time-integrated flow descriptors such as residence time or Lagrangian particle activation, which would capture the temporal nature of thrombosis formation more accurately than instantaneous Eulerian features alone. Building on this, extending the analysis to multiple operating points would clarify how the dominant thrombosis mechanisms change with flow rate and rotational speed, potentially informing both pump design and therapy management. The framework can also be used to compare different macroscopic thrombosis models and to identify mechanistic similarities or inconsistencies among them, which may help improve future model formulations. Progress in these areas will ultimately depend on validation against clinically or experimentally observed thrombus locations, since reliable spatial data are essential for assessing the accuracy and practical relevance of any predictive approach.

But not only rotary blood pumps could benefit from such an approach. The method is also well suited for vascular thrombosis and patient-specific risk assessment, where flow-induced activation plays an important role. In such applications, interpretable models could help identify the flow conditions that promote platelet activation or clot initiation in anatomically complex regions, and they could support the evaluation of treatment strategies by highlighting which hemodynamic features are most strongly associated with disease progression.

## 5. Conclusion

We present a machine learning framework for thrombosis prediction in rotary blood pumps based on LR. Trained on CFD data with labels from a validated macroscopic thrombosis model, the LR classifier achieves high predictive accuracy, generalizes across conditions and devices, and identifies key flow variables driving thrombosis risk. The approach provides a computationally efficient and interpretable alternative to traditional threshold- or transport-based models, with the potential to support both mechanistic understanding and clinical decision making, provided that future work focuses on improving data quality, incorporating temporal flow characteristics, and validating predictions against real-world observations. The code implementing the full machine learning pipeline is openly available at https://doi.org/10.5281/zenodo.17901009 to enable reproducibility and further development by the scientific community.

Thrombosis," *Advanced healthcare materials*, vol. 14, no. 28, e2500436, 2025, doi: 10.1002/adhm.202500436.